\documentclass[sigconf]{acmart}
\AtBeginDocument{%
  }

\copyrightyear{2025}
\acmYear{2025}
\setcopyright{acmlicensed}
\acmConference[MM '25]{Proceedings of the 33rd ACM International Conference on Multimedia}{October 27--31, 2025}{Dublin, Ireland}
\acmBooktitle{Proceedings of the 33rd ACM International Conference on Multimedia (MM '25), October 27--31, 2025, Dublin, Ireland}
\acmDOI{10.1145/3746027.3754698}
\acmISBN{979-8-4007-2035-2/2025/10}

\usepackage{subcaption}

\usepackage{orcidlink}
\usepackage{multirow} 
\usepackage{bbding}
\definecolor{myblue}{RGB}{66,133,244}
\definecolor{mygreen}{RGB}{51,168,83}
\definecolor{myyellow}{RGB}{251,188,3}
\definecolor{myred}{RGB}{234,67,53}
\definecolor{mygrey}{RGB}{95,99,104}
\definecolor{mypup}{RGB}{153,0,204}
\begin{document}
\title{DSDNet: Raw Domain Demoiréing via Dual Color-Space Synergy}

\settopmatter{authorsperrow=4}

\author{Qirui Yang}
\authornotemark[1]
\email{yangqirui@tju.edu.cn}
\affiliation{
  \institution{Tianjin University}
  \city{Tianjin} 
  \country{China} 
}

\author{Fangpu Zhang}
\authornote{Equal contribution.}
\email{zhangfp@tju.edu.cn}
\affiliation{
  \institution{Tianjin University}
  \city{Tianjin} 
  \country{China} 
}

\author{Yeying Jin}
\email{yeyingjin@global.tencent.com}
\affiliation{
  \institution{Tencent}
  \city{Singapore} 
  \country{Singapore} 
}

\author{Qihua Cheng}
\email{chengqihua@microbt.com}
\affiliation{
  \institution{Shenzhen Bit Microelectronics Technology Co., Ltd}
  \city{Shenzhen} 
  \country{China} 
}
\author{Peng-Tao Jiang}
\email{pt.jiang@mail.nankai.edu.cn}
\affiliation{
  \institution{vivo Mobile Communication Co., Ltd}
  \city{Hangzhou} 
  \country{China} 
}

\author{Huanjing Yue}
\authornotemark[2]
\email{huanjing.yue@tju.edu.cn}
\affiliation{
  \institution{Tianjin University}
  \city{Tianjin} 
  \country{China} 
}

\author{Jingyu Yang}
\authornote{Corresponding author.}
\email{yjy@tju.edu.cn}
\affiliation{
  \institution{Tianjin University}
  \city{Tianjin} 
  \country{China}
}
\renewcommand{\shortauthors}{Qirui Yang et al.}

\begin{abstract}
  With the rapid advancement of mobile imaging, capturing screens using smartphones has become a prevalent practice in distance learning and conference recording. However, moiré artifacts, caused by frequency aliasing between display screens and camera sensors, are further amplified by the image signal processing pipeline, leading to severe visual degradation. Existing sRGB domain demoiréing methods struggle with irreversible information loss, while recent two-stage raw domain approaches suffer from information bottlenecks and inference inefficiency. To address these limitations, we propose a single-stage raw domain demoiréing framework, Dual-Stream Demoiréing Network (DSDNet), which leverages the synergy of raw and YCbCr images to remove moiré while preserving luminance and color fidelity. Specifically, to guide luminance correction and moiré removal, we design a raw-to-YCbCr mapping pipeline and introduce the Synergic Attention with Dynamic Modulation (SADM) module. This module enriches the raw-to-sRGB conversion with cross-domain contextual features. Furthermore, to better guide color fidelity, we develop a Luminance-Chrominance Adaptive Transformer (LCAT), which decouples luminance and chrominance representations. Extensive experiments demonstrate that DSDNet outperforms state-of-the-art methods in both visual quality and quantitative evaluation and achieves an inference speed $\mathrm{\textbf{2.4x}}$ faster than the second-best method, highlighting its practical advantages. \textit{We provide an anonymous online demo at \href{https://xxxxxxxxdsdnet.github.io/DSDNet/}{DSDNet}.} 
\end{abstract}

\ccsdesc[500]{Computing methodologies~Computational photography}

\keywords{Image demoir\'eing, Raw domain, Image signal processor}

\begin{teaserfigure}
  \includegraphics[width=0.98\textwidth]{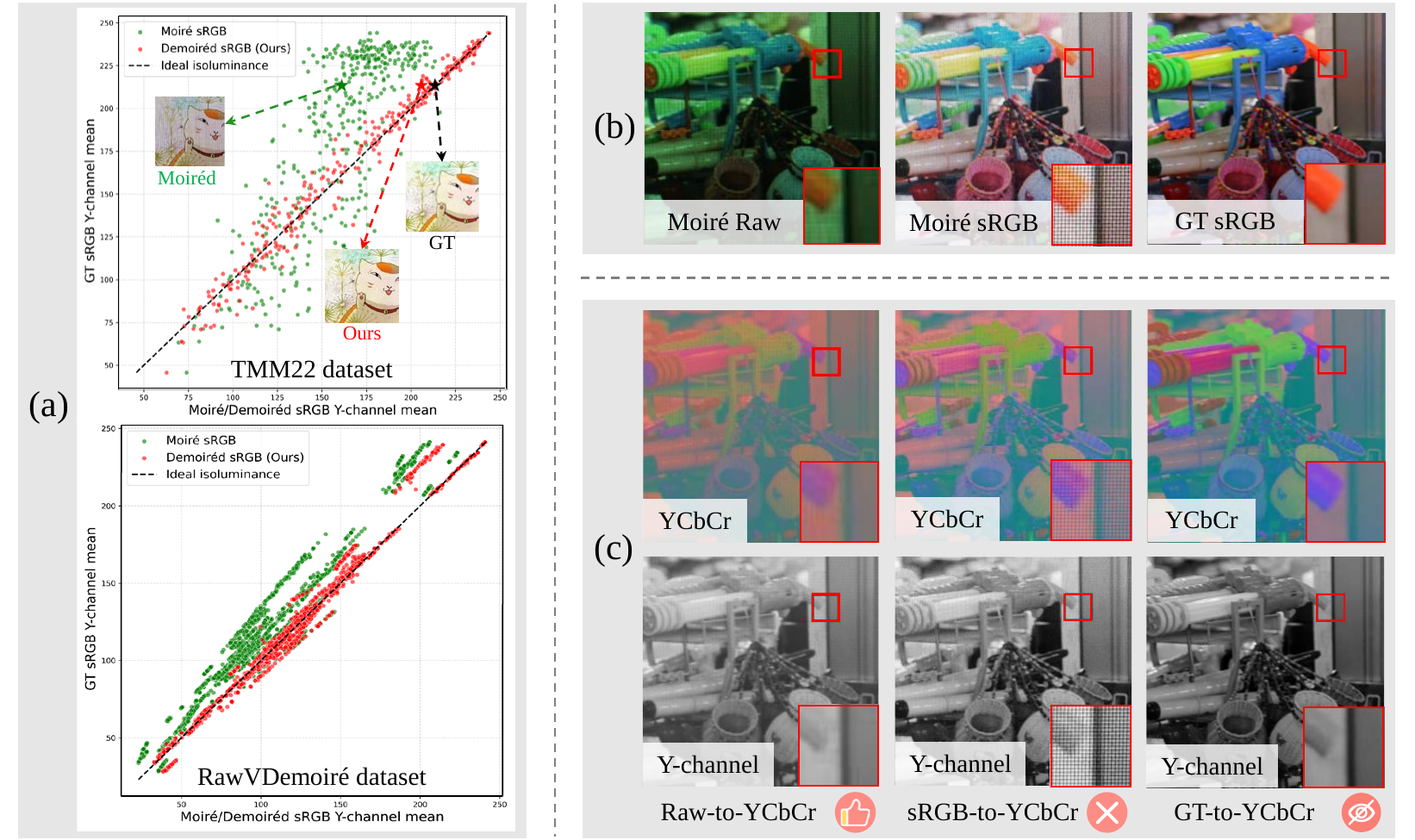}
  \vspace{-0.3cm}
  \caption{Cross color-space observations of moiré artifacts: (a) Moiré images show significant luminance discrepancy (\textcolor{mygreen}{green points}) against the ground truth (GT). Our approach significantly reduces luminance discrepancies in the Y-channel (\textcolor{myred}{red points}). (b) Moiré sRGB images are even more severely degraded by demosaicing, which amplifies moiré artifacts. (c) The Y-channel of raw-to-YCbCr mapping (first column) contains fewer moiré artifacts than the standard sRGB-to-YCbCr mapping (second column), which motivates us to use raw-to-YCbCr images as a prior to guide the luminance and suppress moiré artifacts.}
  \label{fig:main}
\end{teaserfigure}

\maketitle

\begin{figure*}[t]
    \centering
    \includegraphics[width=0.96\linewidth]{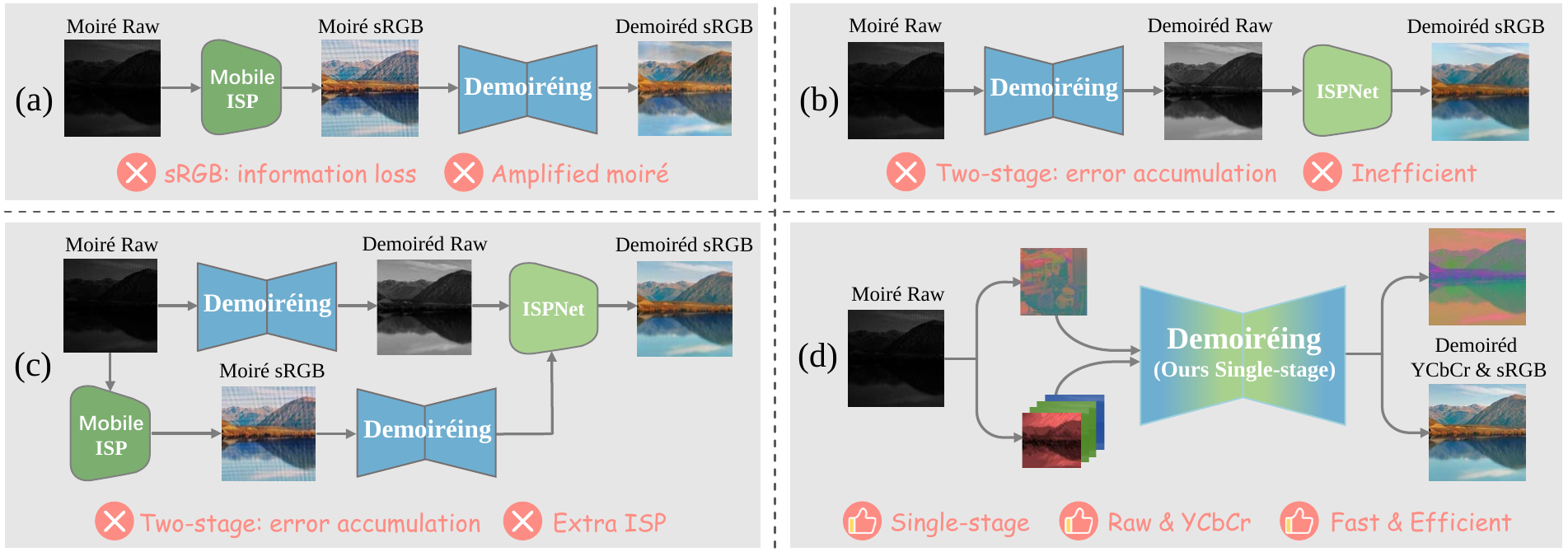}
    \vspace{-0.35cm}
    \caption{Comparison of common paradigms of demoir\'eing models. Mobile ISP denotes the ISP of smartphones. Unlike existing methods, DSDNet inference efficiently uses the single-stage framework and does not require an extra mobile ISP.}
    \label{fig:paradigms}
    \vspace{-0.35cm}
\end{figure*}

\vspace{-0.2cm}
\section{Introduction}
The rapid evolution of mobile imaging technology has made smartphone photography essential for quickly capturing information in various scenarios, including distance learning, conference recording, and technical documentation \cite{liu2015moire, yang2017demoireing}.  
However, a persistent challenge in these settings is the occurrence of moiré artifacts, which are visually disruptive colored stripes. 
These artifacts arise from frequency aliasing between the pixel arrays of digital displays and camera sensors. They are further exacerbated by harmonic amplification introduced by demosaicing in the image signal processing (ISP) pipeline, as shown in Fig. \ref{fig:main}(b). 
This combination of non-linear distortions results in complex, non-stationary moiré patterns that severely degrade image quality and impair the overall visual experience. 
Consequently, effective image demoiréing has become a crucial task in computational photography.

Recent advancements in sRGB-domain demoiréing have been driven by deep learning methods and the availability of large, paired datasets~\cite{cheng2019multi, he2020fhde}.
These methods have introduced innovative strategies such as frequency-domain modeling~\cite{he2019mop}, multiscale feature fusion~\cite{sun2018moire}, and moiré pattern classification~\cite{liu2018demoir}, leading to significant improvements in moiré removal.
However, these sRGB-based methods~\cite{zheng2020image, liu2020wavelet} suffer from two critical limitations. 
First, they often overlook the underlying physical processes of moiré pattern formation, especially the fact that demosaicing in ISP \textbf{amplifies moiré artifacts} \cite{yue2022recaptured}, as shown in Fig. \ref{fig:paradigms}(a).
Second, by passing through the ISP pipeline, images suffer from irreversible information loss, making restoring clean images more challenging.

In response to these limitations, recent research has shifted towards raw domain processing, which preserves the linear characteristics of the sensor signal and better aligns moiré pattern formation with the ISP pipeline. 
Although raw domain demoiréing frameworks have shown significant improvements over traditional sRGB methods, they still face inherent bottlenecks. 
Typically, these frameworks \cite{yue2022recaptured, xu2024image} consist of two stages: the first stage learns the mapping from moiré to clean images, while the second stage learns the mapping from raw to sRGB, as shown in Fig. \ref{fig:paradigms}(b) and (c). 
These methods use cascaded subnetworks, where each stage enhances the output image from the previous one.
With image-level data flow, only images are propagated forward, and each stage relies solely on the result of the prior stage.
This design inherently increases the risk of information loss and \textbf{error accumulation}. 
Additionally, every subnetwork in each stage may incur information loss due to the downsampling operation.
As a result, suboptimal performance arises from the lossy image-level data flow, where errors propagate, accumulate, and amplify across stages, ultimately failing to reconstruct texture details and color fidelity.
Moreover, these two-stage pipelines \cite{song2023real, yue2023recaptured} face challenges in real-time inference and introduce significant computational overhead.
For instance, Xu et al.~\cite{xu2024image} mitigate color cast by adding an sRGB branch, but generating moiré-contaminated sRGB images requires an \textbf{extra ISP} process, which is computationally expensive (see Fig. \ref{fig:paradigms}(c)).
These limitations highlight the need for a more efficient, single-stage raw-domain demoiréing framework that preserves both high fidelity and color accuracy while enabling fast inference.
Such a framework would improve robustness, efficiency, and generalization across diverse imaging devices and real-world scenarios.

Motivated by these challenges, we identified two key observations.
First, due to smartphone auto-exposure adjustments and fluctuations in screen brightness, a noticeable discrepancy exists between the captured moiré input and the true scene radiance (Fig. \ref{fig:main}(a)), primarily affecting the luminance component (Y-channel). 
Second, direct raw-to-YCbCr mapping exhibits fewer moiré artifacts compared to the standard sRGB-to-YCbCr conversion (Fig. \ref{fig:main}(c)).
This is because raw-to-YCbCr mappings are less influenced by ISP-induced harmonic amplification, and chrominance channels (Cb/Cr) are less sensitive to neutral-colored moiré patterns.
These observations inspired us to leverage YCbCr priors to guide the moiréd raw inputs toward clean sRGB outputs, improving structure preservation and color fidelity.
However, directly mapping raw images to the YCbCr color space results in color distortions, necessitating a sophisticated adaptive mechanism to combine the existing raw stream for color mapping.

In light of these findings, we propose the Dual Stream Demoiréing Network (DSDNet), a novel single-stage raw domain demoiréing framework.
DSDNet introduces four key innovations in a unified architecture.
First, we establish a dual stream design that processes inputs in both raw and YCbCr spaces, where the YCbCr stream prioritizes moiré pattern removal and luminance correction, and the raw stream focuses on color mapping. 
Second, we introduce the color mapping block and color guidance block, which extract raw and YCbCr features, respectively.
To effectively bridge these two color spaces, we design the Synergic Attention with Dynamic Modulation (SADM), which facilitates adaptive cross-domain interactions via a dynamic gated fusion strategy based on cross-attention, enabling optimal alignment and complementary information sharing between the streams.
Finally, we introduce the Luminance Chrominance Adaptive Transformer (LCAT), which decouples luminance and chrominance representations using learnable weights to better guide color fidelity restoration.
This framework effectively mitigates nonlinear distortions in the ISP simulation stage, leading to improved perceptual quality.

Our contributions can be summarized as follows: 
\begin{itemize} 
\item We introduce DSDNet, a raw-YCbCr dual stream processing framework for single-stage raw domain image demoiréing, where the YCbCr stream emphasizes moiré removal and the raw stream focuses on color mapping. 
\item We propose that SADM enrich interaction between raw feature and YCbCr feature with cross-domain contextual features through learnable gating mechanisms.
\item We develop the LCAT, which leverages YCbCr priors to guide luminance correction and color fidelity.
\item Our proposed DSDNet outperforms previous methods on several metrics, improving the PSNR on the TMM22 and RawVDemoiré datasets by \textbf{0.47 dB} and \textbf{0.79 dB}, respectively, and achieving an inference speed $\mathrm{\textbf{2.4x}}$ faster than the second-best method.
\end{itemize}

\vspace{-0.1cm}
\section{Related Works}
\label{sec:related}
\subsection{sRGB and Raw Image Demoiréing}
Existing demoiréing methods mainly fall into two categories:

\textbf{sRGB-based demoiréing methods.} To handle the varying scales of moiré patterns, many methods have employed multi-scale architectures~\cite{yang2017textured, cheng2019multi, he2020fhde, liu2020mmdm, wang2021image, niu2023progressive, wang2023coarse, liu2015moire, yang2017demoireing}. 
The field has also advanced through the creation of dedicated real-world datasets~\cite{sun2018moire, yue2020recaptured} and the introduction of priors like edge guidance~\cite{he2019mop}.
Other works have explored the frequency domain using wavelet~\cite{liu2020wavelet} or DCT-based filtering~\cite{zheng2020image, zheng2021learning}. More recently, Transformer-based models like Uformer~\cite{wang2022uformer} and Restormer~\cite{zamir2022restormer} have demonstrated strong performance. The scope has also expanded from static images to video demoiréing~\cite{dai2022video}.

\textbf{Raw-based demoiréing methods.} 
Processing in the raw domain avoids artifacts from the ISP pipeline, such as demosaicing and non-linear transformations. This approach typically preserves more detail and achieves superior performance, as demonstrated in recent works for both image and video demoiréing~\cite{yue2022recaptured, yue2023recaptured, xu2024image}. However, these methods rely on a two-stage pipeline: first removing moiré in the raw domain, then mapping the result to an sRGB image. While this decoupling simplifies the task, it can lead to irreversible information loss, thereby limiting the final image quality.

\vspace{-0.1cm}
\subsection{Learning-Based RAW Image Processing}
Raw image restoration has gained traction due to its advantages over sRGB-based pipelines, spurring progress across several tasks~\cite{liu2019learning, wang2020practical, wei2020physics, yang2025learning, yang2023efficient}. 
Chen et al.~\cite{chen2018learning} built the SID dataset for low-light raw images and introduced a UNet for denoising. 
Building on this, Yang et al.~\cite{yang2025learning} proposed a single-stage network with feature domain adaptation for raw low-light image enhancement.
Beyond low-light enhancement, Zhang et al.~\cite{zhang2019zoom} employed optical zoom to capture LR-HR raw image pairs for super-resolution. 
Zou et al.~\cite{zou2023rawhdr} extended the raw task to HDR reconstruction and noise reduction using multi-exposure bursts. Liu et al. \cite{liu2023joint} built a smartphone-captured HDR dataset covering diverse lighting conditions.
To address multi-task challenges, Qian et al.~\cite{qian2022rethinking} introduced a joint demosaicing, denoising, and super-resolution framework, while Sharif et al.~\cite{a2021beyond} tackled demosaicing and denoising in sensor data.

\begin{figure*}[ht]
    \centering
    \includegraphics[width=0.97\linewidth]{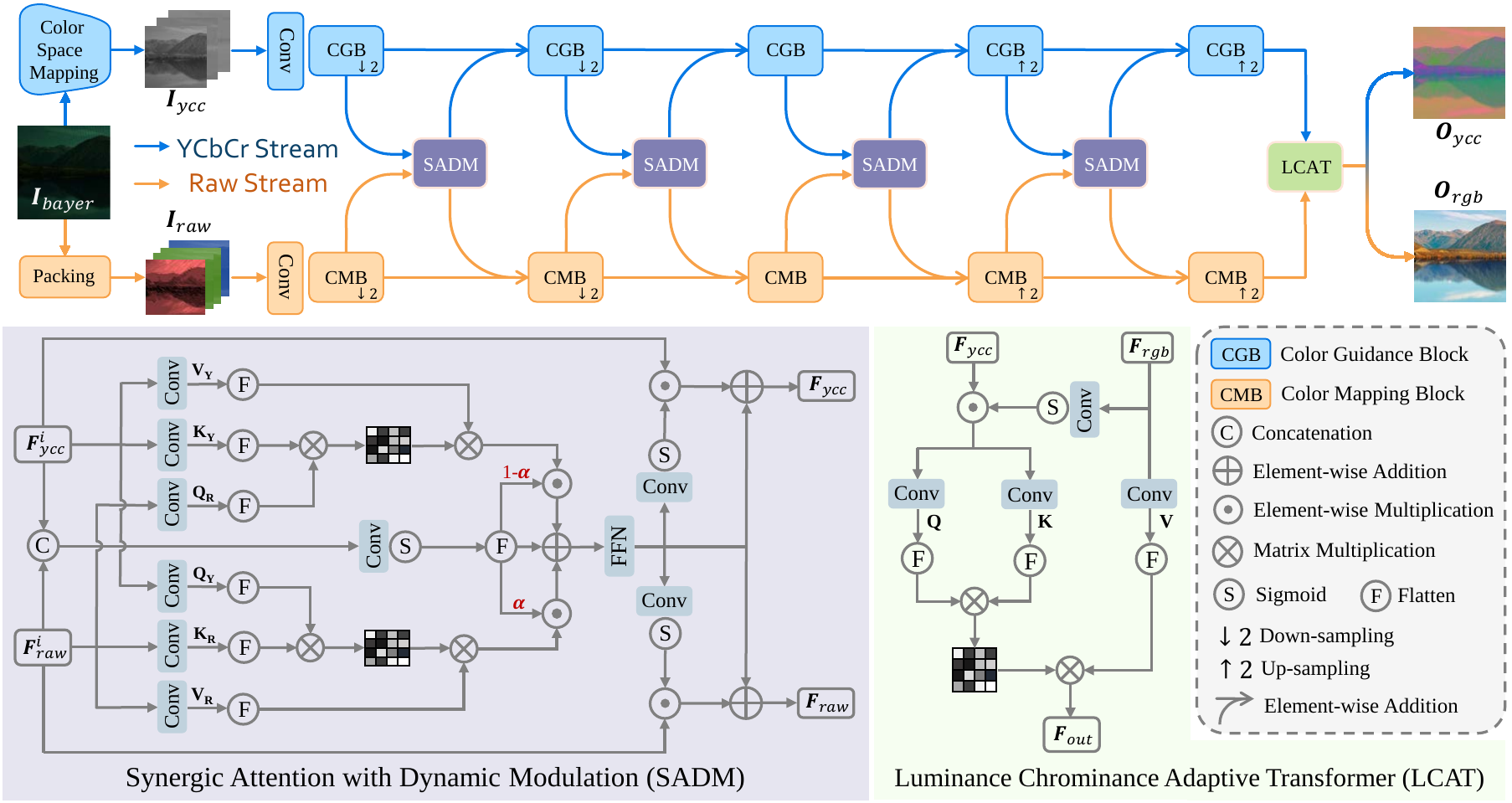}
    \vspace{-0.35cm}
    \caption{The overall pipeline of our DSDNet. It includes the proposed Synergic Attention with Dynamic Modulation (SADM) \textcolor{mypup}{(see purple region)} and Luminance Chrominance Adaptive Transformer (LCAT) \textcolor{mygreen}{(see green region)}. SADM promotes raw features to produce clearer textures through the YCbCr color space, while LCAT significantly enhances the overall visual effect.}
    \label{fig:arch}
    \vspace{-0.35cm}
\end{figure*}

\vspace{-0.1cm}
\section{Method}  
As shown in Fig. \ref{fig:arch}, the proposed DSDNet consists of two complementary streams, raw-to-sRGB and YCbCr-to-YCbCr, that collaboratively restore moiréd raw images.
Given an Bayer input image $\mathbf{I}_\text{bayer} \in \mathbb{R}^{2H \times 2W}$, we first convert it into a 4-channel RGGB format, denoted as $\mathbf{I}_{\text{raw}} \in \mathbb{R}^{4 \times H \times W}$, which serves as the input to the color mapping stream.
Simultaneously, we average the two green channels of $\mathbf{I}_\text{bayer}$ and convert the result into YCbCr space using a standard transformation matrix yielding $\mathbf{I}_{\text{ycc}} \in \mathbb{R}^{3 \times H \times W}$ as the input to the guidance stream.
This dual-stream representation facilitates both moiré removal and color fidelity preservation.
The raw and YCbCr images are independently processed by dedicated encoders and then fed into an SADM module to enrich the representation with cross-domain contextual features.
The integrated features are subsequently routed to two specialized streams tailored for demoiréing and color mapping.
Finally, luminance-chrominance features from the YCbCr stream further guide the RGB reconstruction, improving perceptual quality.

\vspace{-0.25cm}
\subsection{Dual Stream Decoupling}
Raw domain demoiréing poses unique challenges, requiring simultaneous moiré removal, raw-to-sRGB mapping, and luminance correction.
Learning all three tasks jointly within a unified feature space proves inefficient due to their disparate characteristics.
To simplify this issue, we decouple the process via a dual-stream architecture: the raw-to-sRGB stream focuses on color mapping, while the YCbCr stream provides structural and luminance priors that are relatively moiré-resistant.

As shown in Fig. \ref{fig:arch}, the Color Mapping Block (CMB) processes raw features for color mapping, while the Color Guidance Block (CGB) provides guidance for moiré removal and luminance correction. The detailed designs are presented in the Supplementary Material.
The CMB employs a State Space Model (SSM) \cite{guo2024mambair} and a Gated Multilayer Perceptron (GMLP) to model the raw-to-sRGB mapping.
The SSM captures row-wise correlations in a line-by-line fashion, mimicking the physical row scanning of image sensors.
Prior works rely on channel attention to suppress redundancy; however, our analysis reveals that channel redundancy is negligible and that channel attention can disrupt spatial detail preservation critical for demoiréing.
To address this, we introduce GMLP, which jointly captures spatial and channel-wise patterns. 
Specifically, for an input feature $\mathbf{F}_a$, we first expand its channels, split it into two parts, apply depth-wise convolution ($\text{DConv}_{3 \times 3}$) to one half, and fuse them via gated multiplication:
Finally, the two parts are fused via an element-wise gating mechanism. The GMLP process is described as:
\begin{equation} 
{\mathbf{F}_{1}, \mathbf{F}_{2}} = \text{Split}(\text{Linear}(\mathbf{F}_a)),
\end{equation}
\begin{equation} 
\mathbf{F}'_{a} = \text{Linear}(\mathbf{F}_{1} \odot \text{DConv}_{3 \times 3}(\mathbf{F}_{2})), 
\end{equation}
where $\odot$ denotes element-wise multiplication. 
Thus, the CMB output for the raw feature $\mathbf{F}^i_{\text{raw}}$ is given by:
\begin{equation} 
\mathbf{\hat{F}}^i_{\text{raw}} = \text{CMB}(\mathbf{F}^i_{\text{raw}}) = \text{GMLP}(\text{SS2D}(\mathbf{F}^i_{\text{raw}})),
\end{equation} 
where $\text{SS2D}(\cdot)$ represents the 2D state space operator from~\cite{guo2024mambair}.

Meanwhile, the CGB extracts the prior guidance features using dilated channel attention from~\cite{xu2024image}, capturing both global and local dependencies:
\begin{equation} 
\mathbf{\hat{F}}^i_{\text{ycc}} = \text{CGB}(\mathbf{F}^i_{\text{ycc}}). 
\end{equation} 


\vspace{-0.2cm}
\subsection{Synergic Attention with Dynamic Modulation}
To enrich the raw-to-sRGB conversion with cross-domain contextual features, we introduce the SADM module.
SADM enables bidirectional interaction between raw features $\mathbf{\hat{F}}^i_{\text{raw}}$ and YCbCr features $\mathbf{\hat{F}}^i_{\text{ycc}}$ via symmetric attention mechanisms and a dynamic gating fusion pathway. 
Given the encoded raw features $\mathbf{\hat{F}}^i_{\text{raw}} \in \mathbb{R}^{c \times h \times w}$ and YCbCr features $\mathbf{\hat{F}}^i_{\text{ycc}} \in \mathbb{R}^{c \times h \times w}$, SADM first computes cross-domain correlations through symmetric attention pathways. 
In the Raw-to-YCbCr pathway, $\mathbf{\hat{F}}^i_{\text{raw}}$ is projected into key-value pairs ${\mathbf{K}_{\text{raw}}, \mathbf{V}_{\text{raw}}}$ via a $1 \times 1$ convolution ($\text{Conv}_{1\times1}$), while the query $\mathbf{Q}_{\text{ycc}}$ is derived from $\mathbf{\hat{F}}^i_{\text{ycc}}$.
Symmetrically, in the YCbCr-to-Raw pathway, ${\mathbf{K}_{\text{ycc}}, \mathbf{V}_{\text{ycc}}}$ are obtained from $\mathbf{\hat{F}}^i_{\text{ycc}}$ and $\mathbf{Q}_{\text{raw}}$ from $\mathbf{\hat{F}}^i_{\text{raw}}$: 
\begin{equation} 
\mathbf{Q}_{\text{ycc}} = \text{Conv}_{1\times1}(\mathbf{\hat{F}}^i_{\text{ycc}}), \quad {\mathbf{K}_{\text{raw}}, \mathbf{V}_{\text{raw}}} = \text{Split}(\text{Conv}_{1\times1}(\mathbf{\hat{F}}^i_{\text{raw}})), 
\end{equation} 
\begin{equation} 
\mathbf{Q}_{\text{raw}} = \text{Conv}_{1\times1}(\mathbf{\hat{F}}^i_{\text{raw}}), \quad {\mathbf{K}_{\text{ycc}}, \mathbf{V}_{\text{ycc}}} = \text{Split}(\text{Conv}_{1\times1}(\mathbf{\hat{F}}^i_{\text{ycc}})). 
\end{equation}

The attention weights are computed via scaled dot-product attention, followed by softmax normalization, and used to refine the complementary features: 
\begin{equation}
\mathbf{A}_{\text{raw}\to\text{ycc}} = \mathrm{softmax} \left (\frac{\mathbf{\mathbf{Q}_{\text{raw}}\mathbf{K}_{\text{ycc}}}^\top}{\sqrt{d_K}} \times \tau \right)\mathbf{V}_{\text{ycc}},
\end{equation}
\begin{equation}
\mathbf{A}_{\text{ycc}\to\text{raw}} = \mathrm{softmax}\left (\frac{\mathbf{\mathbf{Q}_{\text{ycc}}\mathbf{K}_{\text{raw}}}^\top}{\sqrt{d_K}} \times \tau \right) \mathbf{V}_{\text{raw}},
\end{equation}
where $d_K$ denotes the dimensionality of $\mathbf{K}$, and $\tau$ is a learnable temperature parameter. The outputs $\mathbf{A}_{\text{ycc}\to\text{raw}}$ and $\mathbf{A}_{\text{ycc}\to\text{raw}}$ encapsulate bidirectional contextual dependencies across modalities.

To adaptively merge these pathways, we introduce a dynamic content-aware gating mechanism to learn spatial-wise fusion weights. The concatenated features $[\mathbf{\hat{F}}^i_\text{raw}, \mathbf{\hat{F}}^i_\text{ycc}]$ are processed via two cascaded $1 \times 1$ convolutions with GELU activation:
\begin{equation}
\mathbf{\alpha} = \mathcal{G}([\mathbf{\hat{F}}^i_\text{raw}, \mathbf{\hat{F}}^i_\text{ycc}]) = \sigma\left(\mathcal{C}_2(\text{GELU}(\mathcal{C}_1(\cdot))\right),
\end{equation}
where $\mathcal{G}(\cdot)$ is the dynamic gating function composed of $\mathcal{C}_1$ and $\mathcal{C}_2$ ($1 \times 1$ convolution), followed by GELU and Sigmoid $\sigma$ activations, producing fusion coefficients $\alpha \in [0,1]^{h \times w}$. 
The final fused feature is obtained by blending the two attention outputs:
\begin{equation}
\mathbf{\hat{F}}^i = \alpha \odot \mathbf{A}_{\text{ycc}\to\text{raw}} + (1-\alpha) \odot \mathbf{A}_{\text{raw}\to\text{ycc}}.
\end{equation}

This output is projected to the original dimensionality and further refined by a feed-forward network (FFN) \cite{zamir2022restormer}: 
\begin{equation}
\mathbf{F}^i_{\text{fuse}} = \text{FFN}(\text{Conv}_{1\times1}(\mathbf{\hat{F}}^i)) + \mathbf{\hat{F}}^i.
\end{equation}

To enrich representation, we modulate each stream by channel-spatial recalibration. Specifically, for the raw stream: 
\begin{equation}
\mathbf{M}_\text{raw} = \sigma\left(\text{Conv}_{1\times1}(\text{ReLU}(\text{Conv}{1\times1}(\mathbf{F}^i_{\text{fuse}}))\right)\ ,
\end{equation}
\begin{equation}
\mathbf{F}^{i+1}_{\text{raw}} = \mathbf{\hat{F}}^i_\text{raw} \odot \mathbf{M}_\text{raw} + \mathbf{F}^i_{\text{fuse}}.
\end{equation}
Here, $\mathbf{M}_\text{raw} \in [0, 1]^{c \times h \times w}$ encodes channel-spatial modulation weights, facilitating precise recalibration of raw features while preserving the learned cross-domain semantics.
The YCbCr stream is modulated similarly: 
\begin{equation}
\mathbf{M}_\text{ycc} = \sigma\left(\text{Conv}_{1\times1}(\text{ReLU}(\text{Conv}{1\times1}(\mathbf{F}^i_{\text{fuse}}))\right)\ ,
\end{equation}
\begin{equation}
\mathbf{F}^{i+1}_{\text{ycc}} = \mathbf{\hat{F}}^i_\text{ycc} \odot \mathbf{M}_\text{ycc} + \mathbf{F}^i_{\text{fuse}}.
\end{equation}
Through this modulation mechanism, SADM enables adaptive selection and effective integration of cross-domain features, enriching the raw-to-sRGB conversion with cross-domain contextual features.

\begin{figure*}[t]
    \centering
    \includegraphics[width=0.98\linewidth]{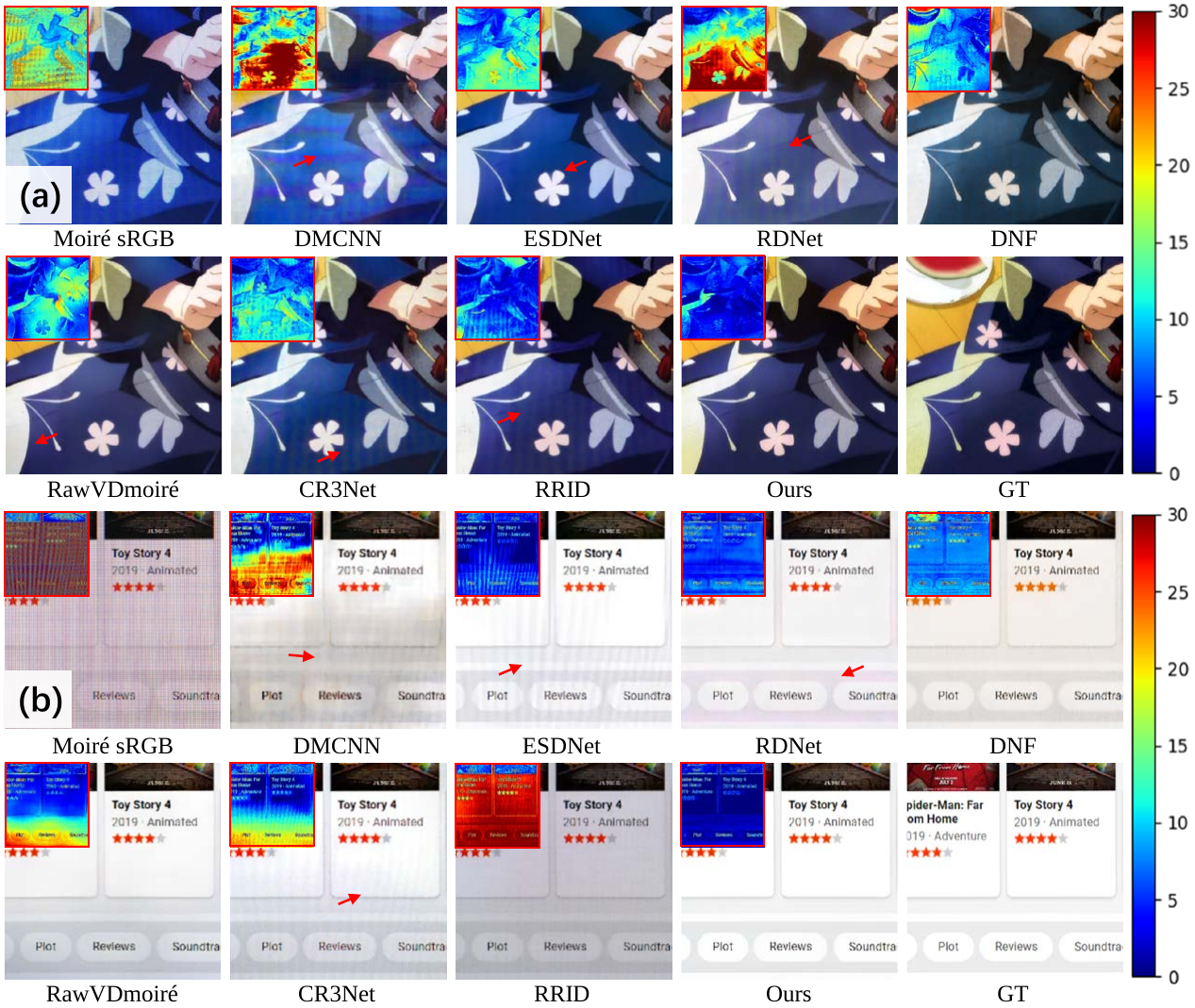}
    \vspace{-0.35cm}
    \caption{Comparison with state-of-the-art demoiréing methods on TMM22 dataset. The error maps in the upper left corner provide clearer insights into performance differences.}
    \label{fig:tmm}
    \vspace{-0.3cm}
\end{figure*}
\begin{table*}[t]
    \centering
    \caption{Quantitative comparison with the state-of-the-art demoiréing and raw image restoration methods on TMM22 dataset \cite{yue2022recaptured} regarding PSNR, SSIM, LPIPS, Y-PSNR, $\Delta E$ and computational complexity. Y-PSNR denotes the PSNR value of the Y channel. Metrics with $\uparrow$ and $\downarrow$ denote higher better and lower better. The best and second results are in \textcolor{red}{red} and  \textcolor{blue}{blue}, respectively. The ”-” symbol indicates that the results are not available.}
    \vspace{-3mm}
    \label{table:quantitativetmm}
        \begin{tabular}{cccccccccc}
        \toprule
        \multirow{3}*{Index} & \multicolumn{9}{c}{Methods}\\
		\cmidrule(r){2-10}
		 ~ & DMCNN & WDNet  & ESDNet & RDNet  & DNF & RawVDmoiré & CR3Net & RRID & DSDNet\\
         ~ & \cite{sun2018moire} & \cite{liu2020wavelet} & \cite{yu2022towards} & \cite{yue2022recaptured} & \cite{jin2023dnf} & \cite{yue2023recaptured} & \cite{song2023real} & \cite{xu2024image} & (Ours)\\
        \midrule
        \# Input type & sRGB & sRGB & sRGB & RAW & RAW & RAW & sRGB+RAW & sRGB+RAW & RAW \\
        \midrule
        PSNR$\uparrow$ & 23.54 & 22.33 & 26.77 & 26.16 & 23.55 & 27.26 & 23.75 &  {\color{blue}\underline{27.88}} & \textbf{{\color{red}28.35}} \\
        SSIM$\uparrow$ & 0.885 & 0.802 &  0.927 & 0.921 & 0.895 & 0.935 & 0.922 & {\color{blue}\underline{0.938}} & \textbf{{\color{red}0.941}} \\
        LPIPS$\downarrow$ & 0.154 & 0.166 & 0.089 & 0.091 & 0.162 & {\color{blue}\underline{0.075}} & 0.102 & 0.079 &  \textbf{{\color{red}0.065}} \\
        Y-PSNR$\uparrow$ & 24.95 & - & 27.80 & 27.21 & 24.60 & 28.39 & 25.27 &  {\color{blue}\underline{28.49}} & \textbf{{\color{red}29.35}} \\
        $\Delta E$$\downarrow$ & 8.66 & - & 7.01 & 6.26 & 10.12 & 5.49 & 6.77 & {\color{blue}\underline{5.35}} &  \textbf{{\color{red}4.71}} \\
        \cmidrule(r){1-10}
        Params (M) & {\color{blue}\underline{1.55}} & 3.36 & 5.93 & 6.04 & \textbf{{\color{red}1.25}} & 5.33 & 2.68 & 2.38 & 2.77 \\
        TFLOPs & 0.102 & {\color{blue}\underline{0.055}}  & 0.141 & 0.161 & \textbf{{\color{red}0.013}} & 0.173 & 0.883 & 0.093 & 0.085  \\
        Inference time (s) & {\color{blue}\underline{0.052}} & 0.284 & 0.115 & 1.094 & 0.070 & 0.182 & 0.058 & 0.089 &  \textbf{{\color{red}0.037}}\\
        \bottomrule
        \end{tabular}
    \vspace{-0.15cm}
\end{table*}

\vspace{-0.2cm}
\subsection{Luminance Chrominance Adaptive Transformer}
To effectively capture the intricate interplay between luminance and chrominance components, we propose the LCAT. Unlike conventional transformers that directly process sRGB features, LCAT dynamically integrates sRGB and YCbCr domains feature representations through an adaptive fusion mechanism. This approach enables better color fidelity by selectively emphasizing relevant chrominance information while maintaining structural consistency.

Given an input feature $\mathbf{F}_{\text{rgb}}$ and corresponding YCbCr representation $\mathbf{F}_{\text{ycc}}$, we first apply layer normalization independently to both representations to ensure stable feature distributions.
To achieve adaptive feature modulation, we introduce a gating mechanism by a convolution block.
This mechanism generates an attention mask that selectively scales chrominance contributions before fusion:
\begin{equation} 
\mathbf{G} = \sigma\left(\text{Conv}_{1\times1}(\mathrm{ReLU}(\text{Conv}_{1\times1}(\mathbf{F}_{\text{rgb}})))\right), 
\end{equation} 
where $\mathbf{G} \in [0, 1]^{c \times h \times w}$  is a learned gating map that modulates the influence of chrominance features. The fusion process is defined as:
\begin{equation} 
\mathbf{F}_\text{{m}} = \mathbf{F}_{\text{ycc}} \odot \mathbf{G} + \mathbf{F}_{\text{rgb}}, 
\end{equation} 
ensuring that only the most relevant chrominance information is integrated into the sRGB representation.

To enhance global feature interactions, LCAT applies self-attention over the fused representation. Specifically, we derive query-key-value representations as follows:
\begin{equation}
\mathbf{Q}, \mathbf{K} = \text{Split}(\text{Conv}_{1\times 1}(\mathbf{F}_\text{m})), \quad 
\mathbf{V} = \text{Conv}_{1\times 1}(\mathbf{F}_\text{rgb}),
\end{equation}
where query ($\mathbf{Q}$) and key ($\mathbf{K}$) are derived from the gated representation to emphasize illumination-adaptive relationships, while values ($\mathbf{V}$) are extracted from the original sRGB domain to maintain color fidelity. The self-attention mechanism is computed as:
\begin{equation}
A = \text{Softmax}\left(\tau_h \frac{\mathbf{Q} \mathbf{K^\top}}{\sqrt{d}}\right)\mathbf{V},
\end{equation}
where $d$ is the dimension of $\mathbf{K}$ and $\tau_h$ denotes the scaling factor. The attended features are then projected back to the feature space and refined through a feedforward network (FFN):
\begin{equation} 
\mathbf{F}_{\text{out}} = \text{FFN}(\text{Norm}(\mathbf{A})) + \mathbf{A}. 
\end{equation}
This formulation allows LCAT to selectively propagate luminance-aware details while ensuring structural coherence across the color spectrum. By leveraging adaptive fusion and self-attention, LCAT effectively bridges sRGB and YCbCr spaces, enhancing image details while preserving natural color balance. 


\section{Experiments}
\label{sec:exp}
\subsection{Experimental Setup}
\textbf{Dataset.}
We evaluate our method on two representative raw domain demoiréing benchmarks: the TMM22 dataset~\cite{yue2022recaptured} and the RawVDemoiré dataset~\cite{yue2023recaptured}. The TMM22 dataset contains 948 pairs of raw and sRGB image pairs (540 for training and 408 for testing), capturing recaptured scenes including natural images, documents, and web pages. These images are obtained using four different smartphone cameras and three display screens of varying sizes. For both training and evaluation, patches of size $256 \times 256$ and $512 \times 512$ are cropped to facilitate model training and fair comparison across methods.
Complementing TMM22, the RawVDemoiré dataset provides a video-based benchmark, consisting of 300 raw and sRGB video pairs (250 for training and 50 for testing), each video containing 60 frames at a resolution of $1280 \times 720$. Similar to TMM22, the acquisition process employs three smartphone cameras and various displays, ensuring device diversity. Following the protocol of Yue et al.~\cite{yue2023recaptured}, we adapt this dataset for image-level demoiréing by extracting individual frames and reorganizing them into 15,000 training and 3,000 testing image pairs.
\begin{figure*}[htb]
    \centering
    \includegraphics[width=0.99\linewidth]{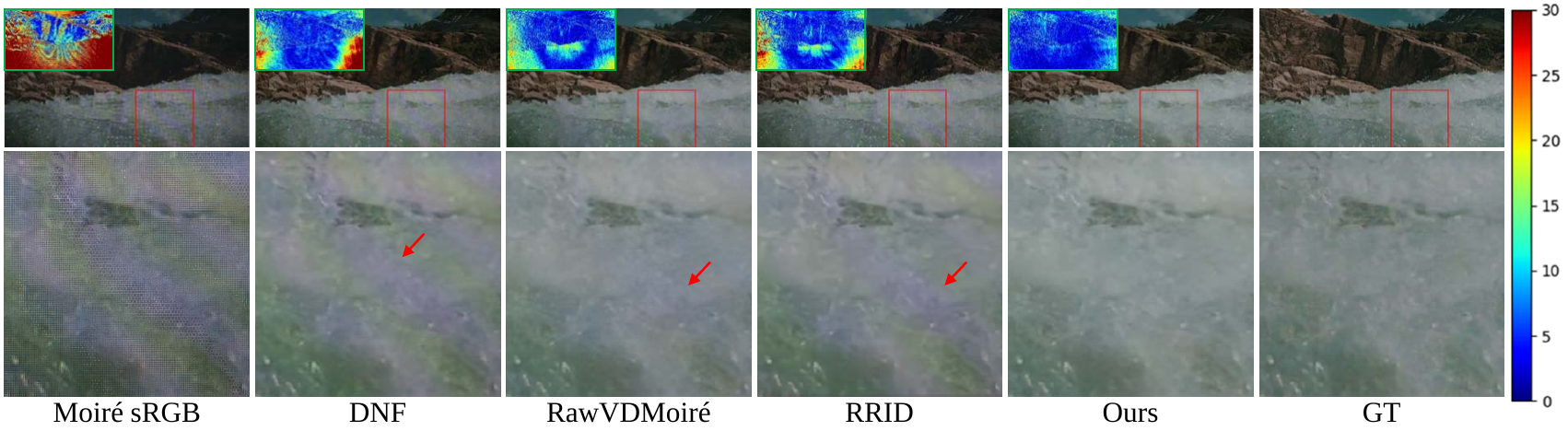}
    \vspace{-0.3cm}
    \caption{Comparison with state-of-the-art demoiréing methods on RawVDemoiré dataset.}
    \label{fig:nips}
    \vspace{-0.3cm}
\end{figure*}

\textbf{Implementation details.}
Our model is implemented using PyTorch and trained on an NVIDIA RTX 3090 GPU. We employ the AdamW optimizer with hyperparameters $\beta_1 = 0.9$ and $\beta_2 = 0.999$ for a total of $4 \times 10^5$ iterations. The learning rate is initialized at $2 \times 10^{-4}$ and gradually reduced to $1 \times 10^{-6}$ via a cosine annealing schedule. For evaluation, we adopt PSNR and SSIM~\cite{wang2004image} computed on the sRGB channel to measure reconstruction fidelity, and PSNR on the Y channel to assess luminance accuracy. To further evaluate perceptual quality, we use LPIPS~\cite{zhang2018unreasonable} and the color difference metric $\Delta E$~\cite{zhang1996spatial}. In addition, model complexity is assessed in terms of parameter count, FLOPs, and inference time.

\subsection{Comparison with State-of-the-Art}
\textbf{Quantitative comparison.} We quantitatively compare the proposed method with state-of-the-art methods, including the sRGB-based methods DMCNN~\cite{sun2018moire}, WDNet~\cite{liu2020wavelet}, ESDNet~\cite{yu2022towards}, and a raw image demoiréing method RDNet~\cite{yue2022recaptured}. We further add comparison with a raw image demoiréing method RRID~\cite{xu2024image} with raw+sRGB input, a video demoiréing method RawVDmoiré~\cite{yue2023recaptured} with raw input, a raw domain low-light enhancement method DNF~\cite{jin2023dnf}, and a dereflection method CR3Net~\cite{song2023real} with paired raw+sRGB data. 
For sRGB-based methods originally designed for 3-channel inputs, we implement architectural adaptations by modifying input channels to accept 4-channel Raw data and inserting bilinear upsampling layers to resolve resolution mismatches. To the RawVDemoiré dataset, we also add several sRGB-based video methods (EDVR~\cite{wang2019edvr}, VDMoiré~\cite{dai2022video}, VRT~\cite{liang2024vrt}, and DTNet~\cite{xu2024direction}) for comprehensive comparison.
To ensure a fair comparison, all baseline implementations are fine-tuned on our training dataset using their officially recommended settings. 
As shown in Tab.~\ref{table:quantitativetmm} and \ref{table:quantitativenips}, our method demonstrates significant advantages on both datasets. Compared to the latest raw demoiréing methods, DSDNet achieves improvements of \textbf{0.47 dB} (PSNR), \textbf{0.003} (SSIM), and \textbf{0.014} (LPIPS) over RRID~\cite{xu2024direction} on the TMM22 dataset. In addition, DSDNet even outperforms the video-based RawVDemoiré~\cite{yue2023recaptured} by \textbf{1.09 dB} in PSNR. On the RawVDemoiré dataset, DSDNet still shows consistency superiority over the second-best method, CR3Net, by \textbf{0.78 dB}.

Notably, DSDNet achieves these results using only a single frame of raw input, eliminating the need for an additional ISP to generate sRGB guidance.
The computational complexity analysis reveals additional strengths: DSDNet achieves the fastest inference time (\textbf{0.037 s/image}) among all compared methods. This efficiency stems from our optimized single-stage architecture, which avoids the computational overhead of multi-stage processing while maintaining reconstruction quality. All metrics are calculated under identical hardware conditions, with the best results highlighted in bold.
\\
\textbf{Qualitative results.} Visual comparisons in Fig. \ref{fig:nips} and \ref{fig:tmm} demonstrate three critical advantages of our method over existing methods. Please zoom in for a better view. 
Previous methods either failed to effectively remove moiré (e.g., RRID, RawVDmoiré, and CR3Net in Figs. \ref{fig:tmm}, and DMCNN, RDNet, CR3Net, and RRID in Fig. \ref{fig:nips}(A)) or suffered from color distortions (e.g., RDNet, DNF, and CR3Net in Fig. \ref{fig:nips}(A), and DNF, ESDNet, CR3Net in Fig. \ref{fig:nips}(B). In some of the results, texture details are lost (e.g., ESDNet, DNF, and EEID in Fig. \ref{fig:nips}(A)). As illustrated in Fig. \ref{fig:tmm} and \ref{fig:nips} shows that by introducing the YCbCr color space from the Raw domain, our method accurately recovers the color and luminance while removing moiré, which ensures accurate color reconstruction and vivid color saturation. More qualitative comparisons can be found in the Supplementary file and webpage comparison.
\begin{table*}[t]
    \centering
    \caption{Quantitative comparison with the state-of-the-art demoiréing methods and RAW image restoration methods on RawVDemoiré dataset \cite{yue2023recaptured} regarding average PSNR, SSIM, LPIPS. The best and second results are in \textcolor{red}{red} and  \textcolor{blue}{blue}, respectively.}
    \vspace{-0.35cm}
    \label{table:quantitativenips}
        \begin{tabular}{ccccccccccccccc}
        \toprule
        \multirow{3}*{Index} & \multicolumn{10}{c}{Methods}\\
		\cmidrule(r){2-11}
		 ~ & RDNet & VDMoiré & EDVR  & VRT  &DTNet & DNF & RawVDmoiré & CR3Net & RRID & DSDNet\\
         ~ & \cite{yu2022towards} & \cite{dai2022video} & \cite{wang2019edvr} & \cite{liang2024vrt} & \cite{xu2024direction} & \cite{jin2023dnf} & \cite{yue2023recaptured} & \cite{song2023real} & \cite{xu2024image} & (Ours)\\
        \midrule
        \# Input type & RAW & RAW & RAW  & RAW  & RAW& RAW & RAW & sRGB+RAW & sRGB+RAW & RAW  \\
        \midrule
        PSNR$\uparrow$ &25.89 & 27.75 & 27.29 & 27.11  & 27.89 & 27.84 & 28.71 & {\color{blue}\underline{28.92}} &  28.37 & \textbf{{\color{red}29.70}} \\
        SSIM$\uparrow$ & 0.894 & 0.912 & 0.906 &  0.903  &0.906 & 0.909 & 0.920 & 0.920 & {\color{blue}\underline{0.921}} & \textbf{{\color{red}0.924}}  \\
        LPIPS$\downarrow$ &0.151 & 0.100 & 0.105 & 0.109  & 0.114 & 0.133 & 0.090 & \textbf{{\color{red}0.085}} & 0.110 & {\color{blue}\underline{0.086}}  \\
        \bottomrule
        \end{tabular}
\vspace{-3mm}
\end{table*}
\vspace{-0.2cm}
\subsection{Ablation Studies}
Our proposed DSDNet integrates three key components: the YCbCr color branch, SADM, and LCAT. To evaluate their individual contributions, we conduct ablation studies on the TMM22 dataset.

\textbf{Effectiveness of specific modules.} To evaluate the effectiveness of each module, we progressively remove the YCbCr branch, SADM, and LCAT, establishing a baseline network that operates solely in the raw domain (w/o YCbCr branch). The results in Tab. \ref{tab:keymodule} show that models using only raw images are limited in performance. When introducing the YCbCr branch (\#2), performance consistently improves, highlighting its effectiveness in color representation.
Comparing \#2 (YCbCr only) with \#3 (YCbCr + SADM, w/o LCAT), we observe a PSNR increase from 27.63 to 27.83 dB, demonstrating that SADM enhances cross-domain interactions and improves reconstruction quality. 
Further incorporating LCAT (\#5, full model) yields a PSNR of 28.35 dB, confirming its role in luminance correction and chrominance-aware reconstruction. 
The full DSDNet model, integrating all three modules, achieves the highest fidelity and perceptual quality, validating the complementary benefits of the YCbCr stream, SADM, and LCAT.

\begin{table}[t]
    \centering
    \caption{Ablation studies of key components.}
    \vspace{-0.3cm}
    \label{tab:keymodule}
    \scalebox{0.95}{
    \begin{tabular}{ccccccc}
        \toprule
        {Variants} & {YCbCr} & {SADM} & {LCAT} & PSNR$\uparrow$ & SSIM$\uparrow$ & LPIPS$\uparrow$\\
        \midrule
        \#1     & \XSolidBrush & \XSolidBrush & \XSolidBrush & 27.29	& 0.9292 & 0.0992\\
        \#2     & \Checkmark & \XSolidBrush & \XSolidBrush & 27.63	& 0.9360  & 0.0735 \\
        \#3     & \Checkmark & \Checkmark & \XSolidBrush & 27.83	& 0.9385 &  0.0677\\
        \#4     & \Checkmark & \XSolidBrush & \Checkmark & 27.84	& 0.9373 & 0.0719\\ 
        \#5     & \Checkmark & \Checkmark & \Checkmark   & \textbf{28.35}  & \textbf{0.9405} & \textbf{0.065}\\
        \bottomrule
\end{tabular}}
\vspace{-0.4cm}
\end{table}
    
\textbf{Effect of color space.} To investigate the impact of color spaces, we conduct experiments using raw, YCbCr, HSV, and YUV representations on the TMM22 dataset. The results, summarized in Tab. \ref{tab:color}, reveal that models restricted to a single color space exhibit limited performance. 
In particular, relying solely on YCbCr leads to the lowest PSNR (26.55 dB) and SSIM (0.9173), indicating its limited ability to handle complex color mapping. 
Combining Raw + HSV performs better than single-space models due to the hue-saturation channel in HSV effectively guiding the mapping from raw to sRGB.
However, the lack of texture cues in HSV reduces its effectiveness in removing moiré patterns. 
In contrast, our full DSDNet model, which leverages YCbCr alongside raw data, achieves the highest performance, demonstrating that color-space transformations significantly enhance both fidelity and perceptual quality. 
These findings validate the effectiveness of our joint raw and YCbCr framework, setting a new benchmark for demoiréing performance.

\begin{table}[t]
\centering
\caption{Quantitative study of different color spaces.}
\vspace{-0.3cm}
\label{tab:color}
\scalebox{1}{
\begin{tabular}{c|cccc}
\toprule
    \textbf{Method} & \textbf{PSNR$\uparrow$} & \textbf{SSIM$\uparrow$} & \textbf{LPIPS$\downarrow$} \\
			\midrule
			Only Raw & 27.29	& 0.9292 & 0.0992\\
			Only YCbCr & 26.55  & 0.9173  & 0.1144  \\
                Raw+HSV  & 28.23 & 0.9399 & 0.0658 \\
                Raw+YUV  & 28.01 & 0.9392 & 0.0663 \\
               \textbf{Ours} & \textbf{28.35}  & \textbf{0.9405} & \textbf{0.065}\\
    \bottomrule
\end{tabular}}
\vspace{-0.3cm}
\end{table}

\subsection{Generalization}
To evaluate the generalization ability of different models, we conduct experiments on the dataset from \cite{yue2022recaptured}, which includes a test set with screen-camera combinations not present in the training set. Specifically, we use these test images—accounting for approximately 25\% of the entire test set—to assess model robustness. As shown in Tab. \ref{tab:gener}, our method achieves the best performance across all metrics, surpassing existing approaches in PSNR, SSIM, and LPIPS. Notably, our method attains a PSNR of 26.261 dB, outperforming RRID by \textbf{1.6 dB} (0.47 dB higher on the same-source dataset) and UHDM by 0.49 dB. Additionally, it achieves the highest SSIM (0.9404) and the lowest LPIPS (0.0673), demonstrating superior perceptual quality and structural preservation. These results highlight the strong generalization capability of our method compared to state-of-the-art approaches.

\begin{table}[htb]
\centering
\vspace{-0.2cm}
\caption{Comparison of the generalization ability of different models by evaluating with the screen-camera combinations that are not included in the training set.}
\vspace{-0.3cm}
\label{tab:gener}
\scalebox{1}{
\begin{tabular}{c|cccc}
\toprule
    \textbf{Method} & \textbf{PSNR$\uparrow$} & \textbf{SSIM$\uparrow$} & \textbf{LPIPS$\downarrow$} \\
			\midrule
			\textbf{FHDe$^2$Net} & 22.840 & 0.8968 & 0.1530 \\
			\textbf{RDNet} & 24.798 & 0.9280 & 0.0846 \\
                \textbf{MBCNN} & 24.726 & 0.9253 & 0.1017 \\
                \textbf{UHDM*} & 25.769 & 0.9310 & 0.0759 \\
			\textbf{RawVDmoiré} & 25.719 & 0.9333 & 0.0727 \\
            \textbf{RRID} & 24.657 & 0.9250 & 0.0985 \\
               \textbf{Ours} & \textbf{26.261} & \textbf{0.9404} & \textbf{0.0673} \\
    \bottomrule
\end{tabular}}
\vspace{-0.4cm}
\end{table}
\section{Conclusion}
In this paper, we propose a novel single-stage image demoiréing network, DSDNet, tailored for raw inputs. 
Our method addresses the limitations of existing methods by avoiding irreversible high-frequency loss and eliminating the inefficiencies of multi-stage pipelines. Extensive experiments demonstrate that DSDNet consistently outperforms state-of-the-art methods in both visual quality and quantitative metrics while achieving the fastest inference speed. The ability to jointly leverage raw information and color-space priors enables robust and efficient moiré removal, making it highly practical for real-world screen capture scenarios. Our work not only advances the field of mobile computational photography but also provides new insights into cross-domain feature learning for image restoration tasks.

\begin{acks}
This work is supported by the National Natural Science Foundation of China under Grants 62231018 and 62472308.
\end{acks}

\bibliographystyle{ACM-Reference-Format}
\bibliography{reference}

\end{document}